\documentclass[]{prime_research}

\usepackage{mathpazo} 
\usepackage{amssymb}
\usepackage{xspace}
\usepackage{mathtools}
\usepackage{tabularx}
\usepackage{enumitem}

\colorlet{seedc}{qyroorange}

\newcommand{\sysname}{SteerUMI\xspace}
\newcommand{\meanTPSgain}{28.6\%}
\newcommand{\meanTPSppgain}{19.9}

\renewcommand{\paragraph}[1]{\vspace{0.1em}\noindent\textbf{#1}}

\crefname{figure}{Figure}{Figures}
\crefname{table}{Table}{Tables}

\title{\sysname: Robot-Free Human-in-the-Loop Post-Training on Distributed UMI Devices}

\author[2,*,\ddagger]{Jiafeng Xu}
\author[1,*]{Qi Li}
\author[1,*]{Xinke Chai}
\author[3]{Xiurui Guo}
\author[2]{Xiaoqi Li}
\author[1]{Xuanming Cheng}
\author[1]{Xiang Li}
\author[2]{Shaowen He}
\author[1]{Hongfei Niu}
\author[1]{Bingye Chen}
\author[1]{Songpeng Zhang}
\author[1]{Yifan Yang}
\author[1]{Qian Zhu}
\author[1]{Ran Cheng}
\author[1,2,\dagger]{Hao Dong}

\affiliation[1]{QYRO}
\affiliation[2]{Peking University}
\affiliation[3]{Beijing Institute of Technology}

\contribution[*]{Equal contribution}
\contribution[\ddagger]{Project lead}
\contribution[\dagger]{Corresponding author.}

\abstract{
Adapting vision-language-action (VLA) models to a deployment requires demonstrations that cover the states the policy actually visits, yet robot-executed human-in-the-loop methods rely on costly physical policy rollouts.
We present \textbf{\sysname}, a cloud-edge framework for robot-free, policy-guided post-training using distributed handheld Universal Manipulation Interface (UMI) devices.
During collection, we take timestamp-aligned human--policy disagreement as a proxy for policy-specific challenging regions, referred to operationally as out-of-distribution (OOD) regions.
Asynchronous evaluation provides feedback without pausing human collection, while augmented reality (AR) guidance encourages locally compatible corrective demonstrations. The collected data support automated DAgger-style iterative aggregation and policy updates.
We evaluate five real-world dual-arm manipulation tasks using 2-8 devices per experiment within a deployment of over 50 UMI devices. With matched per-round episode counts, \sysname achieves a \textbf{\meanTPSgain} mean relative improvement in final-round Task Progress Score (TPS) over conventional UMI collection, averaged across the five tasks.
On stack cubes, it narrows the gap to real-robot HG-DAgger to 3.0 TPS points after three rounds, and with one additional round it achieves a higher final TPS in less collection time, reducing per-round collection time by \textbf{33.3\%} in our single-unit comparison.
}

\checkdata[Project Page]{\url{https://qyrotech.github.io/steerumi-website/}}
\correspondence{
    \email{hao.dong@pku.edu.cn}\textcolor{qyrolink}{,}
    \email{hao.dong@qiyuan-robot.com}
}

\begin{document}
\maketitle

\section{Introduction}
\label{sec:introduction}
\begin{figure*}[t]
\centering
\includegraphics[width=1.0\textwidth]{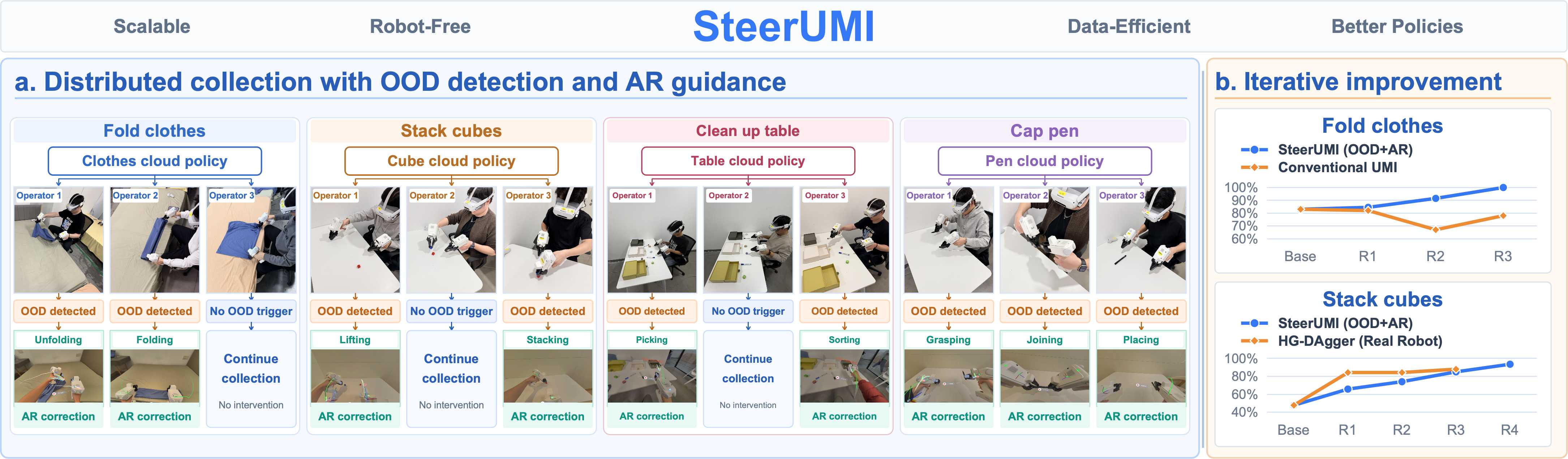}
\caption{
\textbf{\sysname: Robot-free, policy-guided VLA post-training.} (a) Distributed UMI collection with OOD feedback and AR-guided corrections. (b) Iterative TPS (Task Progress Score) improvement compared with conventional UMI collection and real-robot HG-DAgger under same per-round episodes. Conventional UMI collection collects SFT demonstrations without policy feedback, real-robot HG-DAgger collects corrective demonstrations directly through physical robot rollouts, and \sysname collects policy-guided demonstrations with OOD detection and AR guidance.}
\vspace{-0.2cm}
\label{fig:teaser}
\end{figure*}
Vision-language-action (VLA) models have demonstrated promising generalization capabilities in robotic manipulation~\cite{brohan2023rt2}\cite{cheang2025gr3}\cite{kim2024openvla}\cite{nvidia2025gr00tn1}\cite{intelligence2025pi05}, yet adapting them to real-world deployments remains challenging~\cite{kim2025openvloft}\cite{intelligence2025pistar06}.
Effective post-training requires identifying states where the current policy struggles and collecting targeted human demonstrations.
However, existing human-in-the-loop approaches typically rely on repeated policy rollouts and human interventions on physical robots~\cite{hoque2021lazydagger}\cite{kelly2019hgdagger}\cite{liu2023robot}\cite{luo2024rlif}\cite{mandlekar2020human}, incurring substantial hardware costs and limiting scalability.
Recent efforts have explored handheld Universal Manipulation Interface (UMI) devices for robot-free demonstration collection and policy adaptation~\cite{chi2024umi}\cite{yang2026vista}\cite{zhaxizhuoma2025fastumi}. Nevertheless, these efforts have largely focused on local, manually coordinated workflows, leaving distributed data collection and continuous policy improvement largely unexplored.

However, scaling robot-free data collection that keeps improving the policy introduces two challenges:
(1) identifying policy weaknesses without physical robot rollouts, and (2) collecting corrective demonstrations without unnecessarily shifting the existing policy distribution.
To address these challenges, we introduce \textbf{\sysname}, a cloud-edge framework for scalable robot-free human-in-the-loop post-training.
Starting from a pretrained VLA policy, distributed UMI operators continuously collect demonstrations while streaming observations to the cloud for policy inference.
Predicted trajectories are returned to edge devices for policy evaluation and demonstration guidance.
Collected data is then aggregated through an automated online DAgger-style pipeline for continuous policy updates and redeployment.

Specifically, as shown in \Cref{fig:teaser}, to enable effective policy evaluation and demonstration guidance, \sysname introduces two key designs. First, we propose an \textit{asynchronous trajectory-based OOD detector} that identifies potential policy weaknesses by comparing predicted trajectories with timestamp-aligned human demonstrations. Unlike conventional OOD detection that primarily relies on model uncertainty or isolated observations, our approach explicitly measures trajectory-level behavioral disagreement between the current policy and human demonstrations, enabling policy-aware data selection without physical robot rollouts. High-disagreement segments are automatically labeled for targeted post-training, while asynchronous inference allows uninterrupted data collection across distributed devices, even under network delays, thereby maintaining high data collection efficiency.
Second, we introduce \textit{AR-guided policy-aligned demonstration}, which visualizes policy-predicted trajectories through a head-mounted display. Rather than treating human demonstrations as independent supervision, our approach explicitly incorporates the current policy's predicted behaviors into the data collection process, bridging conventional offline UMI collection and policy-driven DAgger. Operators can follow predictions while correcting inappropriate ones, enabling the collection of informative demonstrations that not only remain close to the existing policy distribution but also introduce targeted out-of-distribution (OOD) corrections within the vicinity of that distribution. Together, these designs facilitate policy improvement while reducing unnecessary behavioral shifts, without requiring physical robot rollouts.


We evaluate \sysname on five real-world long-horizon and precision dual-arm manipulation tasks involving over 50 UMI devices.
Our real-time guided data collection approach consistently improves policy performance across all tasks,
achieving a \textbf{\meanTPSgain} mean relative improvement in final-round Task Progress Score (TPS) over conventional UMI collection, averaged across the five tasks.
Compared with DAgger on physical robots on stack cubes, \sysname closes the per-round gap to 3.0 TPS points by the third round and outperforms it with less collection time, as each round takes \textbf{33.3\%} less time to collect.
Our main contributions are summarized as follows:
\begin{itemize}
    \item We introduce \textbf{\sysname}, a scalable cloud-edge framework for robot-free human-in-the-loop VLA post-training, enabling distributed UMI data collection and continuous policy improvement.

    \item We propose an AR-guided policy-aligned demonstration method that encourages the collection of OOD demonstrations within the neighborhood of the original policy distribution, reducing the risk of excessive distributional shifts that could cause model collapse.

    \item We provide a systematic real-world evaluation across five dual-arm manipulation tasks, assessing demonstration effectiveness against conventional UMI collection, collection efficiency against real-robot HG-DAgger, and the contribution of AR guidance through ablation.
\end{itemize}

\section{Related Work}
\label{sec:related_work}

\subsection{Robot-Free Demonstrations for VLA Training}

Robot-free demonstration collection~\cite{kareer2024egomimic}\cite{wang2024dexcap} provides a scalable alternative to costly robot teleoperation~\cite{cheng2024opentelevision}\cite{fu2024mobilealoha}.
ACT-1~\cite{sundayteam2025act1} from Sunday Robotics trains a robot foundation model without any robot data, using a glove that shares the robot hand's geometry and sensor layout to capture human skills and convert them into robot-equivalent training data.
UMI~\cite{chi2024umi} enables the transfer of handheld human demonstrations to robot policies, while FastUMI~\cite{zhaxizhuoma2025fastumi} simplifies hardware setup and data acquisition.
Recent works further explore the use of UMI data for VLA training.
VISTA~\cite{yang2026vista} addresses visual and physical mismatches between UMI demonstrations and robot execution, while HiFi-UMI~\cite{simpleai2026hifiumi} demonstrates effective VLA post-training using high-fidelity robot-free data.
Building on these advances, \sysname focuses on scalable and continuous policy improvement through distributed UMI data collection.

\subsection{Policy-Guided Data Collection and Interactive Learning}

Interactive imitation learning improves policies by collecting targeted demonstrations in challenging states~\cite{laskey2017dart}\cite{menda2019ensembledagger}.
DAgger~\cite{ross2011dagger} introduces iterative dataset aggregation, while ThriftyDAgger~\cite{hoque2022thriftydagger} reduces human supervision by identifying novel or risky states.
For robot-free collection, EgoGuide~\cite{xu2026egoguide} provides online data coverage guidance, while HIL-UMI~\cite{han2026hilumi} leverages policy disagreement and advantage estimation to guide UMI demonstrations.
\sysname instead combines timestamp-aligned, step-wise disagreement evaluation with asynchronous feedback and AR-guided collection in a distributed cloud-edge loop.

\subsection{Distributed and Continual Robot Learning Systems}

Continual robot learning requires efficient coordination between data collection, policy training, and deployment.
Early systems scale robot learning by running distributed data collection and training across many robots~\cite{kalashnikov2018qtopt}, and recent VLA models keep improving by learning from their own deployment experience~\cite{intelligence2025pistar06}.
Fleet-DAgger~\cite{hoque2023fleetdagger} explores scalable human supervision across robot fleets, AutoRT~\cite{ahn2024autort} orchestrates large robot fleets with foundation models, while HIL-SERL~\cite{luo2024hilserl} integrates human corrections with online reinforcement learning.
More recently, SOP~\cite{pan2026sop} enables distributed VLA post-training through centralized cloud learning and asynchronous policy deployment.
Unlike these robot-dependent systems, \sysname establishes a distributed cloud-edge framework that integrates robot-free UMI data collection with automated online DAgger-style iterative aggregation, enabling continuous policy improvement without physical robot rollouts.

\section{Method}
\label{sec:method}

\begin{figure*}[t]
\vspace{-6pt}
\centering
\includegraphics[width=0.85\textwidth]{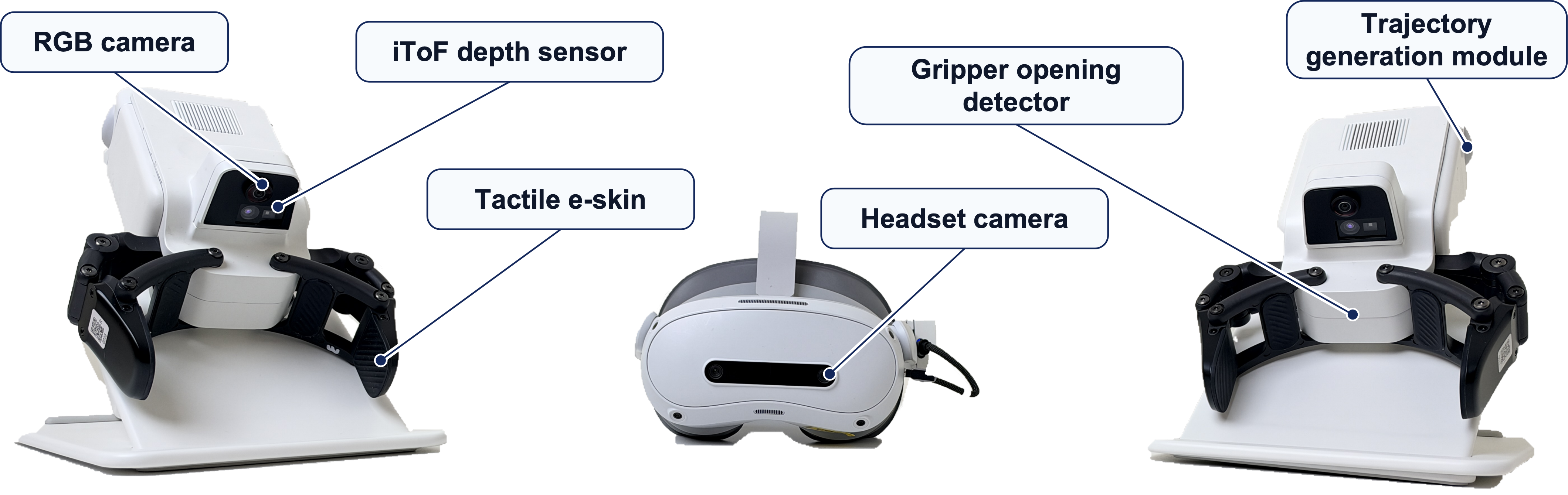}
\caption{\textbf{Sensor layout of the hardware.} The hardware configuration is identical for both the left and right hands.}
\label{fig:hardware}
\end{figure*}

\subsection{\sysname Framework Overview}
\subsubsection{Hardware Design}
The data collection system comprises two handheld UMI devices and a PICO 4 Ultra Enterprise headset.

The system is built around an RK3576 System-on-Chip (SoC) and uses wireless communication.
A hot-swappable 21700 lithium-ion battery supports more than two hours of operation per charge.
Each UMI device incorporates a fisheye RGB camera with a field of view of $200^\circ \times 155^\circ$ and a maximum resolution of $1920 \times 1536$ pixels, capturing the workspace at 30~Hz.
An integrated indirect time-of-flight (iToF) depth camera provides a measurement accuracy of 1~mm within a range of 20~cm.

The PICO headset provides pose tracking and visualizes predicted trajectories for the operator.
It simultaneously records a first-person view of the workspace at a resolution of $1024 \times 768$ pixels and a frame rate of 30~fps.
Pose tracking follows an outside-in approach, achieving millimeter-level absolute trajectory error.
All onboard sensors are synchronized at the hardware level, with a synchronization error below $5\,\mu\mathrm{s}$.
Over 48~hours of continuous operation, the relative timing drift between the two handheld devices remains within one frame.

To reduce the gap between UMI-collected demonstrations and data collected on the physical robot, the gripper adopts the same mechanical design as the OmniPicker~3~\cite{agilink_op3}.
It provides a maximum opening of 118~mm and an effective gripping length of 88.78~mm.
The fingertips are additionally equipped with electronic skin to capture tactile information, with a force measurement range of 0.3--30~N and a sensitivity of 0.1~N.
The specified accuracy is $\leq 1\%$ of full scale (FS) for loads of 0--200~g and $\leq 10\%$ for loads of 200--2500~g.

\subsubsection{Problem Setup}
We consider the post-training of a policy $\pi_\theta$ that has acquired initial manipulation capabilities from a demonstration dataset $\mathcal{D}_0$. While \sysname is applicable to general manipulation policies, we instantiate $\pi_\theta$ as a vision-language-action (VLA) model $\pi_{0.5}$~\cite{intelligence2025pi05}. Given a visual observation $I_t$, a robot state $s_t$, and a language instruction $\ell$, the policy predicts an action chunk of horizon $H$:
\begin{equation}
    \hat{A}_{t:t+H} = \pi_\theta(I_t, s_t, \ell).
\end{equation}
Starting from the initial policy $\pi_{\theta_0}$, our goal is to continuously improve its manipulation performance through iterative DAgger-style data collection and policy updates.
Following the DAgger paradigm, we iteratively identify weaknesses in the current policy, collect corrective human demonstrations, and aggregate them with existing data to update the policy.
Unlike conventional approaches that rely on physical robot rollouts, \sysname performs the entire post-training data collection process using handheld UMI devices, enabling scalable policy improvement without robot execution.

\begin{figure*}[t]
\centering
\includegraphics[width=1.0\textwidth]{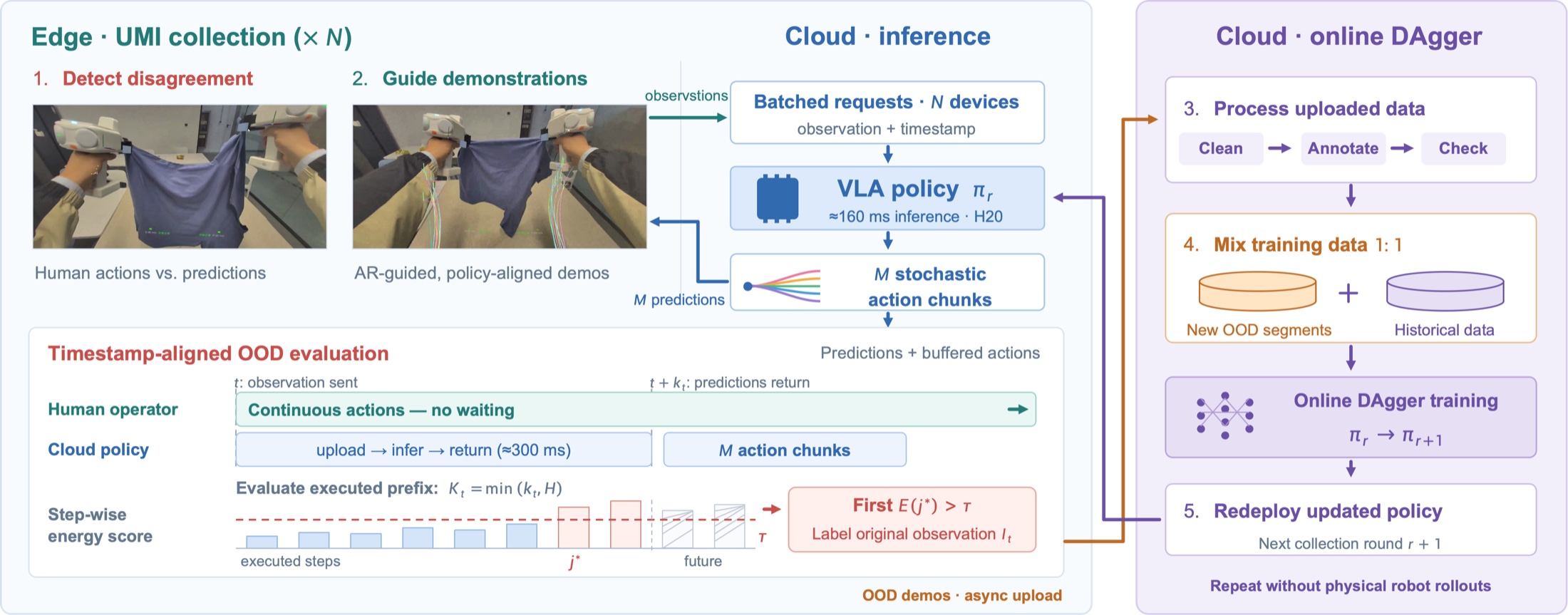}
\caption{\textbf{The \sysname framework.} At prediction return, OOD evaluation compares timestamp-aligned predictions with executed human actions; hatched bars denote unobserved future steps. AR guides corrective collection without pausing human manipulation. The cloud mixes uploaded OOD segments 1:1 with historical data to update the policy.}
\vspace{-0.2cm}
\label{fig:method}
\end{figure*}

\subsubsection{Closed-Loop Workflow}
As illustrated in Figure.~\ref{fig:method}, \sysname establishes a cloud-edge learning loop connecting $N$ distributed UMI devices to a centralized policy server. During each collection round $r$, operators continuously perform manipulation demonstrations while their devices asynchronously stream observations to the cloud. The current policy $\pi_{\theta_r}$ predicts future action trajectories, which are returned to the corresponding edge devices for two complementary purposes: OOD detection and AR-guided demonstration collection.

For OOD detection (refer to Sec.~\ref{sec:ood}), each predicted trajectory is aligned with the human-executed trajectory using its original observation timestamp. Their disagreement is evaluated to identify potentially challenging states, and high-disagreement segments are automatically assigned OOD labels for subsequent policy improvement training.
Meanwhile, predicted trajectories are visualized through AR headsets to guide operators in collecting policy-aligned demonstrations that complement the existing data distribution by covering underrepresented regions (refer to Sec.~\ref{sec:ar-guided}). Importantly, human manipulation proceeds independently of cloud inference, avoiding interruptions caused by communication or inference latency.

Collected demonstrations and their timestamp-aligned OOD labels are asynchronously uploaded to the cloud and processed by an automated streaming pipeline.
\emph{Cleaning} removes episodes with abnormal end-effector velocity, acceleration, or jitter, or discontinuous state recordings;
\emph{annotation} identifies valid, non-stationary frames;
and \emph{quality inspection} checks the recorded video streams.

Once sufficient new data has been collected, the policy is updated using the accumulated dataset (refer to Sec.~\ref{sec:dagger}) and redeployed as $\pi_{\theta_{r+1}}$. This process repeats, forming a continuous loop of policy evaluation, targeted demonstration collection, and iterative improvement.

\subsection{Asynchronous Trajectory-Based OOD Detection}
\label{sec:ood}
During each DAgger-style data collection round, UMI operators initially perform task demonstrations freely, while the cloud policy continuously predicts actions based on their observations.
In this section, we elaborate on our trajectory-based OOD detection mechanism and asynchronous evaluation pipeline, which jointly enable targeted identification of informative demonstrations without interrupting continuous data collection.
\subsubsection{Cloud-Assisted OOD Detection}
To identify informative demonstrations for subsequent policy improvement, we detect policy-specific OOD regions by comparing policy predictions with human-executed actions.
We use human--policy disagreement as a proxy for policy-specific challenging regions.
Since multiple actions may be valid under the same observation, a single prediction may not reliably characterize the policy's behavior.
We therefore sample $M$ stochastic action chunks of horizon $H$ to approximate the policy's predictive distribution:
\begin{equation}
    \hat{A}_t^{(m)} \sim
    \pi_{\theta_r}(\cdot \mid I_t, s_t,\ell),
    \quad m=1,\ldots,M,
\end{equation}
where $\pi_{\theta_r}$ denotes the current policy at iteration $r$.
Meanwhile, the operator continuously executes actions
$A_t=\{a_{t+j}\}_{j=0}^{k_t-1}$,
where $k_t$ denotes the number of executed steps before the predictions return.

Rather than evaluating disagreement over an entire action chunk, we perform step-wise OOD detection over the overlapping horizon $K_t=\min(k_t,H)$.
Following the Energy Score used in HIL-UMI~\cite{han2026hilumi}, we measure the disagreement between each human action and the policy's predictive distribution as follows:
\begin{equation}
\begin{aligned}
E_{t,j} ={}&
\frac{1}{M}\sum_{m=1}^{M}
d(\hat{a}_{t+j}^{(m)},a_{t+j}) \\
&-\frac{1}{2M(M-1)}
\sum_{m\neq n}
d(\hat{a}_{t+j}^{(m)},\hat{a}_{t+j}^{(n)}),
\end{aligned}
\label{eq:d}
\end{equation}
where $d(\cdot,\cdot)$ measures the weighted discrepancy between two actions. Specifically, it combines the Euclidean distances of end-effector positions and gripper states with the geodesic distance of end-effector orientations on $SO(3)$.
The first term in Eq.~\ref{eq:d} measures the average human-policy discrepancy, while the second accounts for the diversity of policy predictions, preventing valid multimodal behaviors from being unnecessarily penalized.

After the predictions return, we evaluate the buffered human actions and identify the earliest threshold-exceeding step as follows:
\begin{equation}
    j^*=\min\{j\in[0,K_t-1]:E_{t,j}>\tau\},
\end{equation}
where $\tau$ is a predefined threshold.
If no action exceeds the threshold, the system proceeds to evaluate subsequent predictions.
Otherwise, $I_t$ is labeled as OOD and $t$ becomes the episode's OOD onset. Further OOD evaluation is disabled, while the operator completes the task. Feedback is issued after prediction return, not at the historical step $t+j^*$; evaluating the executed prefix avoids waiting for the full horizon.

\subsubsection{Asynchronous and Non-Blocking Evaluation}
To support scalable OOD detection across distributed devices, \sysname adopts an asynchronous batched inference mechanism.
Given concurrent observations from $N$ UMI devices, the cloud server aggregates them into a batch and performs parallel policy inference.
In our implementation, the end-to-end image-to-action latency on an NVIDIA H20, measured from transmitting an image observation to receiving the corresponding action prediction, is approximately $300$ ms, including about $160$ ms of policy inference, while operators continuously collect demonstrations at $30$ Hz. This corresponds to approximately $9$ human action steps during each inference round, allowing step-wise comparison within the overlapping prediction horizon when $H>9$.

Each request carries a device identifier and observation timestamp $t$. At prediction return $t+k_t$, the edge evaluates the timestamp-aligned human actions from its local buffer.
Importantly, human manipulation is independent of cloud inference and communication, eliminating inference-induced interruptions during data collection.
Together, asynchronous execution and batched inference enable continuous, scalable OOD detection across distributed UMI devices without requiring physical robot rollouts.

\subsection{AR-Guided Policy-Aligned Demonstration}
\label{sec:ar-guided}
Having identified discrepancies between policy predictions and human demonstrations, we introduce an AR-guided collection mechanism to improve subsequent demonstrations. Our goal is to expand the local coverage of the current policy's behavior distribution while avoiding excessive deviations from previously acquired behaviors, thereby facilitating stable and continuous policy improvement.

\subsubsection{AR Visualization of Policy Trajectories}

Given $M$ action trajectories predicted by the current policy, we extract their corresponding 3D end-effector trajectories and project them onto the 2D display plane of the head-mounted device (\emph{e.g.,} PICO 4 VR headset). Specifically, using the calibrated intrinsic matrix $K_h$ and the world-to-camera transformation $T_{C_h\leftarrow W}$ of the head-mounted camera, we define the projection operator as
\begin{equation}
    \mathcal{P}_h(\mathbf{p}_W)
    =
    \Pi\left(
    K_h
    \begin{bmatrix}
    I_3 & \mathbf{0}
    \end{bmatrix}
    T_{C_h\leftarrow W}
    \begin{bmatrix}
    \mathbf{p}_W\\
    1
    \end{bmatrix}
    \right),
\end{equation}
where $\mathbf{p}_W\in\mathbb{R}^3$ denotes a 3D point in the world coordinate frame, and $\Pi(\cdot)$ performs homogeneous coordinate normalization. Each predicted waypoint is then projected as
\begin{equation}
    \mathbf{u}_{t+j}^{(m)}
    =
    \mathcal{P}_h\left(
    \hat{\mathbf{p}}_{t+j,W}^{(m)}
    \right),
\end{equation}
where $\hat{\mathbf{p}}_{t+j,W}^{(m)}$ denotes the predicted 3D end-effector position at step $t+j$ in the $m$-th trajectory. The resulting trajectory bundle is overlaid onto the operator's egocentric view as shown in Figure.~\ref{fig:method}, providing an intuitive visualization of the policy's predicted behaviors.

\subsubsection{Policy-Aligned and OOD-Guided Collection}

Based on the AR visualization, we guide operators with two complementary collection strategies.

\textbf{First, policy-aligned local exploration.}
Operators are encouraged to collect demonstrations around the displayed trajectories rather than strictly imitate a single prediction. Such local variations expand the coverage of the existing policy behavior distribution while remaining close to its predicted actions, reducing unnecessary distribution shifts during iterative post-training. This provides additional behavioral coverage around the current policy while preserving consistency with previously acquired behaviors.

\textbf{Second, OOD-guided targeted collection.}
For each detected OOD event, we retrieve the recorded 3D end-effector position at the original observation time $t$, denoted as $\mathbf{p}_W^{\mathrm{OOD}}$. The marker identifies the state from which the flagged trajectory was predicted, not the position at $t+j^*$ or at prediction return. This position is projected onto the AR display as a spatial marker:
\begin{equation}
    \mathbf{u}^{\mathrm{OOD}}
    =
    \mathcal{P}_h(\mathbf{p}_W^{\mathrm{OOD}}).
\end{equation}
These markers highlight observation-time locations associated with predicted trajectories that were subsequently flagged by human--policy disagreement. During subsequent collection rounds, operators are encouraged to revisit these marked locations and collect additional demonstrations around them, providing targeted corrective data for previously identified policy weaknesses. By explicitly indicating where the policy struggled, this spatial guidance directs human collection efforts toward informative regions, improving data acquisition efficiency compared with unguided exploration.

Together, these two strategies enable \sysname to expand the local coverage of existing policy behaviors while collecting targeted corrections in previously identified OOD regions, facilitating continuous policy improvement while preserving previously acquired capabilities.

\subsection{Distributed Online DAgger Post-Training}
\label{sec:dagger}

Following each DAgger data collection round, we describe how the collected demonstrations are transferred back to the cloud and incorporated into model training.

During data collection, the network bandwidth of each UMI device is prioritized for real-time OOD detection. Therefore, collected demonstrations are first stored locally and then asynchronously uploaded to the cloud, together with their corresponding OOD labels and the version of the policy used for detection (\emph{e.g.,} the $r^{th}$ round). In the cloud, data from all devices pass through an automated pipeline for data cleaning, annotation, and quality inspection. Further details of the data pipeline are provided in Appendix~\ref{app:data_pipeline}.

Post-training uses only the suffix from the OOD-labeled observation time $t$ to episode end for each newly collected valid OOD episode. Once the OOD data generated under the previous policy version $\pi_{\theta_r}$ reaches a predefined threshold, the cloud server automatically combines it with historical data at a 1:1 ratio for continued policy training. The historical dataset consists of the initial SFT data and DAgger data collected in previous rounds. Specifically, we randomly sample historical segments to match the volume of newly collected OOD data, ensuring an equal data contribution from both sources during training. This enables the policy to continuously learn from newly collected demonstrations while mitigating forgetting of previously acquired behavioral distributions.

\begin{figure*}[t]
\centering
\includegraphics[width=1.0\textwidth]{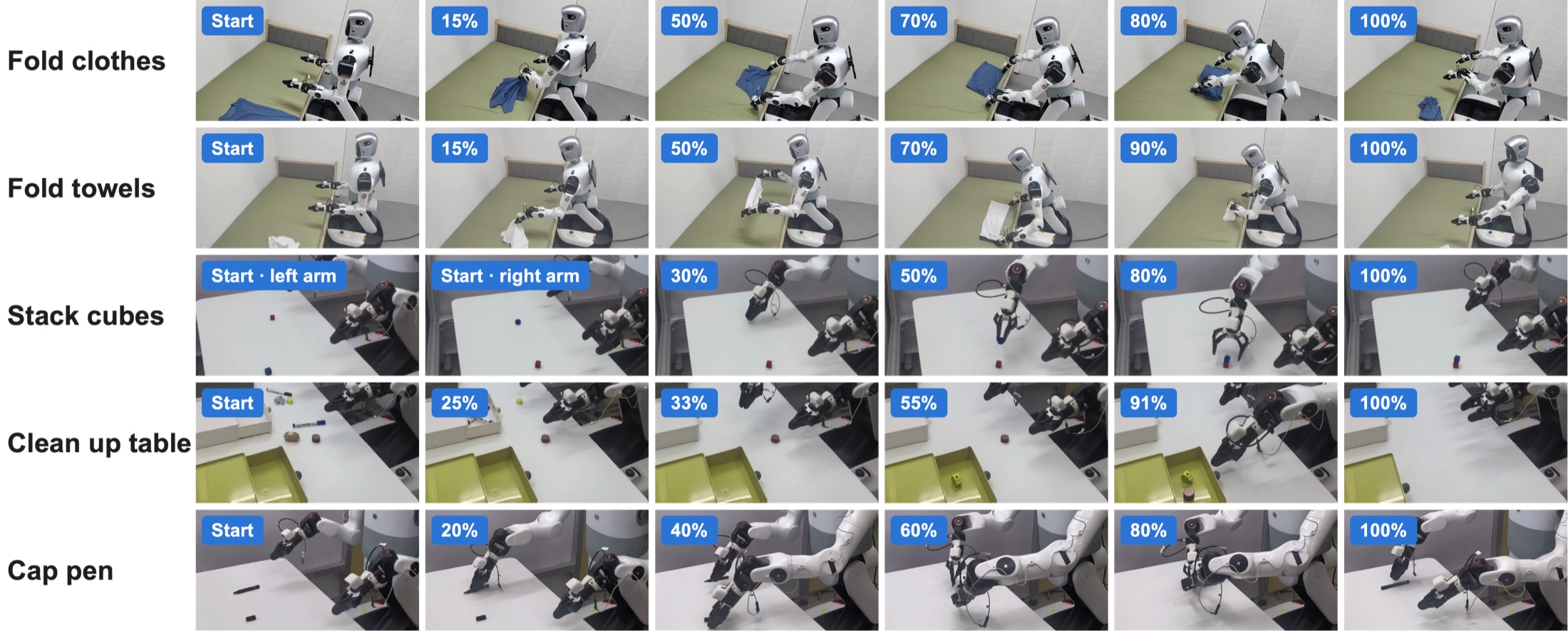}
\caption{\textbf{Five real-world dual-arm tasks.} Each row shows key frames of one task, ordered by progress; the badge on each frame is the Task Progress Score of the milestone it reaches.}
\label{fig:tasks}
\end{figure*}

In each post-training round, the model is trained for a fixed six epochs. The checkpoint from the final (sixth) epoch is used as the round's final model and deployed as the updated policy $\pi_{\theta_{r+1}}$ for OOD detection in the next collection round. The system then automatically publishes a new data collection task to the operators, completing the new round of data collection, cloud-based training, and policy redeployment.

\section{Experiments}
\label{sec:experiments}



We evaluate \sysname on real-world dual-arm tasks to answer two questions:
(1) Given the same number of collected episodes per round, does OOD-guided UMI collection yield a stronger policy than conventional UMI collection?
(2) How does \sysname compare with DAgger on real robots in performance and in iteration efficiency?
Here, conventional UMI collection refers to standard offline data collection for supervised fine-tuning (SFT), where operators gather demonstrations independently of the current policy, without OOD detection or AR guidance.

\subsection{Hardware Setup}
Over 50 UMI devices are deployed in total for data collection and policy iteration.
Each experiment was equipped with 2--8 UMI devices depending on the collection difficulty to ensure that the iteration progress of different tasks was similar.
The gripper configuration and camera selection for each UMI unit were matched to those of the QYRO H2 robot used in the real-world deployment, thereby minimizing the transfer gap.
Although our method is designed to be applicable across different robotic embodiments, we adopt this specific hardware configuration for evaluation.

\subsection{Comparison with Conventional UMI Collection}
We test five real-world dual-arm manipulation tasks (\cref{fig:tasks}).
\textbf{Fold clothes}: the robot moves a shirt to the folding area, flattens the shirt and folds it neatly, and places it back in the storage area.
\textbf{Fold towels}: For single-layered, regularly shaped items like towels, the same folding process used for clothes is applied.
\textbf{Stack cubes}: the robot picks up a cube, moves it above a target cube, places it on top, and moves the arm away.
\textbf{Clean up table}: the robot sorts three groups of items into three boxes, putting two items into each box and then closing it.
\textbf{Cap pen}: the robot picks up a pen and a cap, holds them horizontally, joins them, and places the capped pen on the table.
Together they cover deformable-object manipulation, fine-grained manipulation, bimanual coordination, and long-horizon execution.


For each task, we first collect demonstrations using conventional UMI and train a base model by supervised fine-tuning (SFT). Starting from the same task-specific base model, we compare two iterative post-training groups: one continues with conventional UMI collection, while the other collects demonstrations under OOD detection and AR guidance. For OOD detection, the current policy samples $M{=}10$ stochastic action chunks at each step.
The initial SFT dataset sizes are chosen to establish basic task proficiency before post-training, allowing us to evaluate improvements to an existing policy. Dataset sizes, OOD thresholds, and per-round quotas of 50--120 episodes are detailed in Appendix~B. The thresholds and quotas are set according to task difficulty, while training and detector settings are provided in Appendix~C.

The episode quotas match the number of demonstrations used for training, rather than the number of acquisition attempts or training frames. For \sysname, only valid OOD episodes count toward the quota, and training uses the suffix from the OOD-labeled observation time $t$ to the episode end. Conventional UMI collection contributes the same number of training episodes but uses complete trajectories.

\begin{figure*}[t]
\centering
\includegraphics[width=1.0\textwidth]{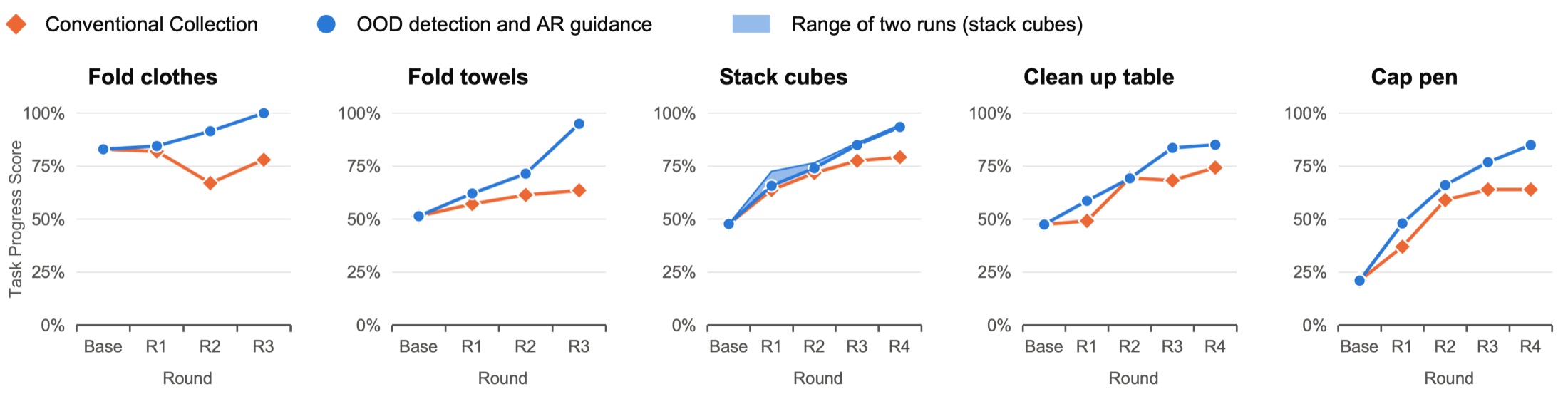}
\caption{\textbf{Task Progress Score over iteration rounds.} Starting from the same base model (Round 0), we iterate with conventional UMI collection (orange) or with OOD detection and AR guidance (blue) on five real-world tasks.}
\label{fig:dagger_exp}

\end{figure*}

\paragraph{Results.} We report the TPS (Task Progress Score): each task is divided into milestones with fixed scores (milestones and scores for each task are listed in Appendix~A).
\Cref{fig:dagger_exp} shows that \sysname improves steadily over rounds on all five tasks, whereas conventional UMI collection does not: on fold clothes, its TPS fluctuates and ends at 0.78, below the base model (0.83), while \sysname reaches 1.00. On fold towels, conventional UMI collection improves from 0.51 to 0.64 over three rounds, while \sysname reaches 0.95. On stack cubes, conventional UMI collection reaches a TPS of 0.79 while \sysname reaches 0.94 after four rounds; on cap pen, \sysname goes from 0.21 to 0.85 after four rounds, versus 0.64 after four rounds of conventional UMI collection.

We compute the relative TPS improvement for each task at its final reported round and then average these gains equally across the five tasks:
\begin{equation}
    \overline{\Delta}_{\mathrm{rel}}
    = \frac{100}{5}\sum_{q=1}^{5}
    \frac{T_q^{\mathrm{ours}}-T_q^{\mathrm{UMI}}}{T_q^{\mathrm{UMI}}},
    \label{eq:mean-relative-tps}
\end{equation}
where $T_q^{\mathrm{ours}}$ and $T_q^{\mathrm{UMI}}$ are the final-round TPS values for task $q$ under \sysname and conventional UMI collection, respectively. This gives a mean relative improvement of \textbf{\meanTPSgain} and a mean absolute gain of  \textbf{\meanTPSppgain} percentage points.

\paragraph{Run-to-run variance.}
To evaluate the robustness and consistency of our approach, we conduct a variance analysis on task \textit{stack cubes} across two independent runs. Specifically, both runs start from the same base policy $\pi_{\theta_0}$, while the DAgger data collected at each round is independently gathered, \emph{e.g.}, by different human operators. We observe that our approach consistently improves performance over the base policy across both runs, with highly similar learning curves and improvement trends as shown by the shaded region in \cref{fig:dagger_exp}. These results demonstrate the stability and robustness of our DAgger data collection pipeline against variations in human demonstrations.

\subsection{Comparison with Real-Robot DAgger}
We compare \sysname against HG-DAgger~\cite{kelly2019hgdagger}, a conventional real-robot DAgger approach, on the stack cubes task.
HG-DAgger follows the standard DAgger paradigm: the current policy is deployed for autonomous rollouts, and a human operator intervenes to correct the robot's behavior when it approaches failure.
Specifically, at each iteration, we deploy the current policy on an NVIDIA RTX 5090 for real-time inference and robot control. When the robot is likely to fail, an operator takes over through teleoperation to correct the ongoing execution and complete the task, while the system records demonstrations collected during these interventions for subsequent policy updates.

Both methods use 100 training episodes per round with the same data-processing and training workflow. For \sysname, the 100 episodes are valid OOD demonstrations. Collection time also includes attempts that do not trigger OOD. We compare one robot for HG-DAgger against one UMI device for \sysname.


\begin{figure}[!htbp]
\centering
\includegraphics[width=0.65\textwidth]{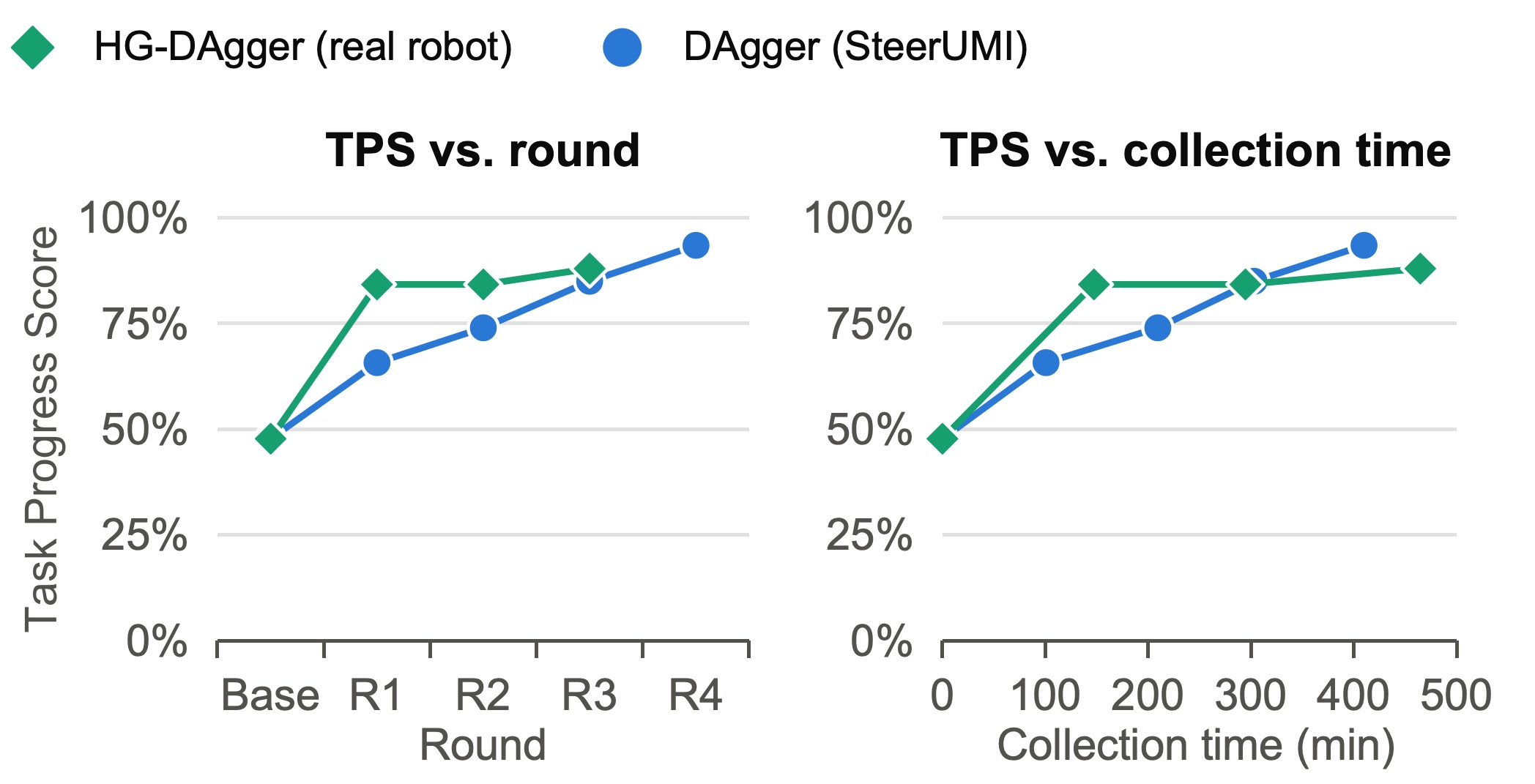}
\captionsetup{width=0.8\textwidth}
\caption{\textbf{\sysname and real-robot HG-DAgger on stack cubes.}
Task Progress Score per round (left) and per collection time (right); both methods collect 100 trajectories per round.}
\label{fig:hg_dagger}
\end{figure}

\subsubsection{Task Performance}


\Cref{fig:hg_dagger} compares the two methods. After the same number of rounds, HG-DAgger is slightly ahead: its TPS is 0.84 after the first round and 0.88 after the third, versus 0.66 and 0.85 TPS for \sysname.
This is expected, because HG-DAgger collects corrective demonstrations directly during policy rollouts on the target robot, so the data are more closely aligned with the policy's state-action distribution.
Under the same per-round training-episode quota, HG-DAgger therefore provides a strong reference for the learning effectiveness of DAgger demonstrations. However, each \sysname round takes less time to collect (about 100\,min vs.\ 150\,min for HG-DAgger), so we run one additional round: with less total collection time (about 410\,min for \sysname vs.\ 460\,min for HG-DAgger), \sysname reaches a TPS of 0.94 versus 0.88 for HG-DAgger.
Round-start setup adds 10\,min per method in total to the 400/450\,min acquisition times, yielding 410/460\,min.
These results show that HG-DAgger learns faster per round, but the gap narrows from 18.5 TPS points after the first round to 3.0 after the third; with one additional round, \sysname surpasses it with less collection time.

\subsubsection{Iteration Efficiency}


We compare acquisition time on the stack-cubes task for a quota of 100 training episodes per round, excluding the separately reported round-start setup time. With a single real robot, collecting one trajectory takes 1.5\,min on average, including starting the rollout, collecting the intervention data, and resetting the robot and the scene, so 100 trajectories take 150\,min.
With a single UMI device, one collection attempt takes 30\,s on average and only one in every 2 attempts triggers OOD, so producing 100 valid OOD trajectories takes $100 \times 2 \times 30\,\mathrm{s} \approx 100$\,min.
Using a single unit of each, \sysname therefore reduces the time per round by about 33.3\% (100 vs.\ 150\,min). Scaling up \sysname with additional UMI devices is also far cheaper than scaling up with additional real robots.

In our local-inference implementation, the latest model is deployed on the robot side for inference, so every iteration transfers about 8\,GB of model weights between cloud and robot (taking $\pi_{0.5}$ as an example), which adds further time. \sysname deploys the model in the cloud throughout the iteration, so the deployment time is negligible.

\section{Ablation Study}
\label{sec:ablation}


To isolate the effect of AR-guided policy-aligned collection, we start from the same model for each task and collect the next round of Dagger data under two settings: with and without AR guidance. In the without-AR setting, operators independently decide how to correct OOD behaviors based on their own judgment, without visual access to the policy's predicted trajectories or OOD locations. In contrast, with AR guidance, operators collect demonstrations around the visualized policy trajectories and are guided to revisit previously identified OOD locations for targeted corrections. Both settings use the same OOD detection mechanism and collect the same amount of data per round. For Fold Clothes, we start from the model after the first DAgger round and train for a second round; for Stack Cubes, we start from the model after the second round and train for a third round.

\begin{table}[t]
\centering
\small
\setlength{\tabcolsep}{3pt}
\caption{Ablation on AR-guided policy-aligned collection. ``Start'' is the model each round
starts from.}
\label{tab:ab_align}
\begin{tabular}{llcccc}
\toprule
& & \multicolumn{2}{c}{Start} & \multicolumn{2}{c}{After} \\
\cmidrule(lr){3-4}\cmidrule(lr){5-6}
Task & Method & SR & TPS & SR & TPS \\
\midrule
\raisebox{-0.6ex}[0pt][0pt]{Fold clothes}
  & w/o AR guidance        & 80.0 & 84.5 & 40.0          & 67.5 \\
  & w/ AR guidance (Ours)  & 80.0 & 84.5 & \textbf{90.0} & \textbf{91.5} \\
\midrule
\raisebox{-0.6ex}[0pt][0pt]{Stack cubes}
  & w/o AR guidance        & 40.0 & 74.0 & 15.0          & 70.0 \\
  & w/ AR guidance (Ours)  & 40.0 & 74.0 & \textbf{55.0}  & \textbf{85.0} \\
\bottomrule
\end{tabular}
\vspace{-0.2cm}
\end{table}

\Cref{tab:ab_align} reports the results. Without AR guidance, operators determine how to correct OOD behaviors at their own discretion, with no guarantee that corrections remain within the neighborhood of the current policy's distribution. Such unconstrained corrections can introduce substantial distribution shifts, making it difficult for the model to learn from new demonstrations while preserving previously acquired capabilities. Consequently, performance can even fall below that of the initial model: on Fold Clothes, the success rate drops from 80.0\% to 40.0\%, while on Stack Cubes, it decreases from 40.0\% to 15.0\%.

In contrast, AR guidance encourages local exploration around the policy's predictions, keeping newly collected demonstrations close to its existing behavior distribution. 
This helps the model learn corrective behaviors while limiting disruption to previously acquired capabilities.
As a result, Fold Clothes achieves a success rate of 90.0\% and a TPS of 91.5\%, while Stack Cubes improves to 55\% success rate and 85\% TPS. These results highlight the importance of distribution-aware data collection: corrective demonstrations are most beneficial when they provide new information while remaining compatible with the policy's existing behavior distribution.

\section{Conclusion}
\label{sec:conclusion}

We present \sysname, a cloud-edge system that enables efficient post-training of vision-language-action models through a human-in-the-loop data collection process, without requiring the participation of real-robots.
Asynchronous trajectory-based OOD detection flags policy weaknesses without physical robot rollouts, and AR-guided demonstrations keep the corrections near the current policy distribution; distributed UMI devices share one cloud policy and an automated DAgger-style pipeline that closes the loop from collection to redeployment.
On five real-world dual-arm tasks, \sysname consistently improves the TPS over conventional UMI collection.
Compared with real-robot HG-DAgger on stack cubes, it narrows the per-round gap to 3.0 TPS points by the third round and outperforms it with less collection time.
In future work, we plan to extend the collection paradigm in one direction: letting the model give improvement suggestions to operators in advance, rather than only flagging failures after they occur. We believe robot-free, policy-guided human feedback is a promising direction for scalable continual adaptation of manipulation policies.

\clearpage

\bibliographystyle{unsrtnat}
\bibliography{reference}

\clearpage
\appendix
\begingroup
\captionsetup[figure]{position=bottom,skip=6pt}
\captionsetup[table]{position=top,skip=6pt}
\setlength{\floatsep}{10pt plus 2pt minus 2pt}
\setlength{\textfloatsep}{12pt plus 2pt minus 3pt}
\setlength{\intextsep}{10pt plus 2pt minus 2pt}

\section{Distributed UMI Data Collection}
\label{ap:distributed_collection}
Figure~\ref{fig:supp_distributed_collection} shows part of the data collection process for the five tasks investigated in the main paper.
In each subfigure, multiple operators perform demonstrations for the same task while using handheld UMI devices and wearing AR headsets.

\begin{figure}[!htbp]
\centering
\includegraphics[width=\linewidth]{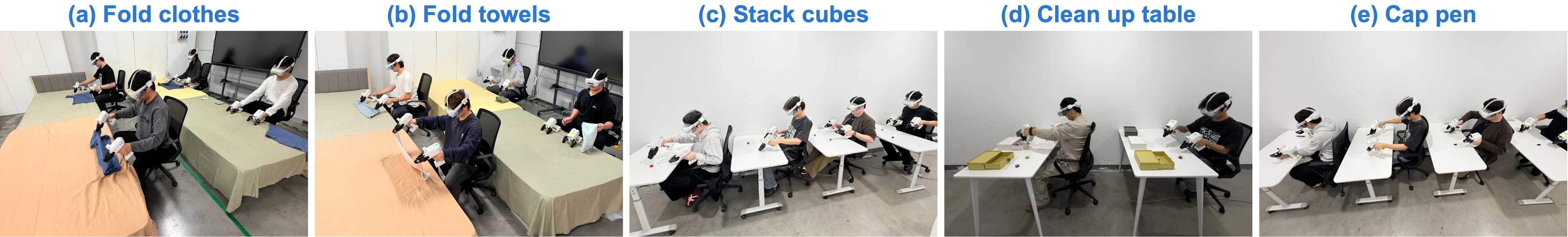}
\caption{\textbf{Representative distributed UMI collection sessions.}
(a) Fold clothes, (b) fold towels, (c) stack cubes, (d) clean up table, and
(e) cap pen. Each photograph is from a separate task session. The panels
illustrate the collection setting and operator equipment, rather than measure
cloud concurrency or collection throughput.}
\label{fig:supp_distributed_collection}
\end{figure}

\FloatBarrier
\section{First-Person Examples of OOD Feedback and AR Guidance}
\label{ap:ood_ar_examples}
Figure~\ref{fig:supp_ood_ar_examples} presents first-person sequences for
all five tasks: fold clothes, fold towels, stack cubes, clean up table, and
cap pen. Each column shows one task through four stages: before OOD feedback,
the first recorded frame displaying the original \texttt{OOD Detected} prompt,
AR guidance during continued manipulation, and the task-completion view.
Before the alert, the displayed trajectories are white. During AR guidance,
the operator's manipulation approaches the colored policy-predicted
trajectories visualized through AR. The images are recorded frames; no
synthetic trajectory overlays or energy-score curves have been added.

\begin{figure}[!htbp]
\centering
\includegraphics[width=\linewidth]{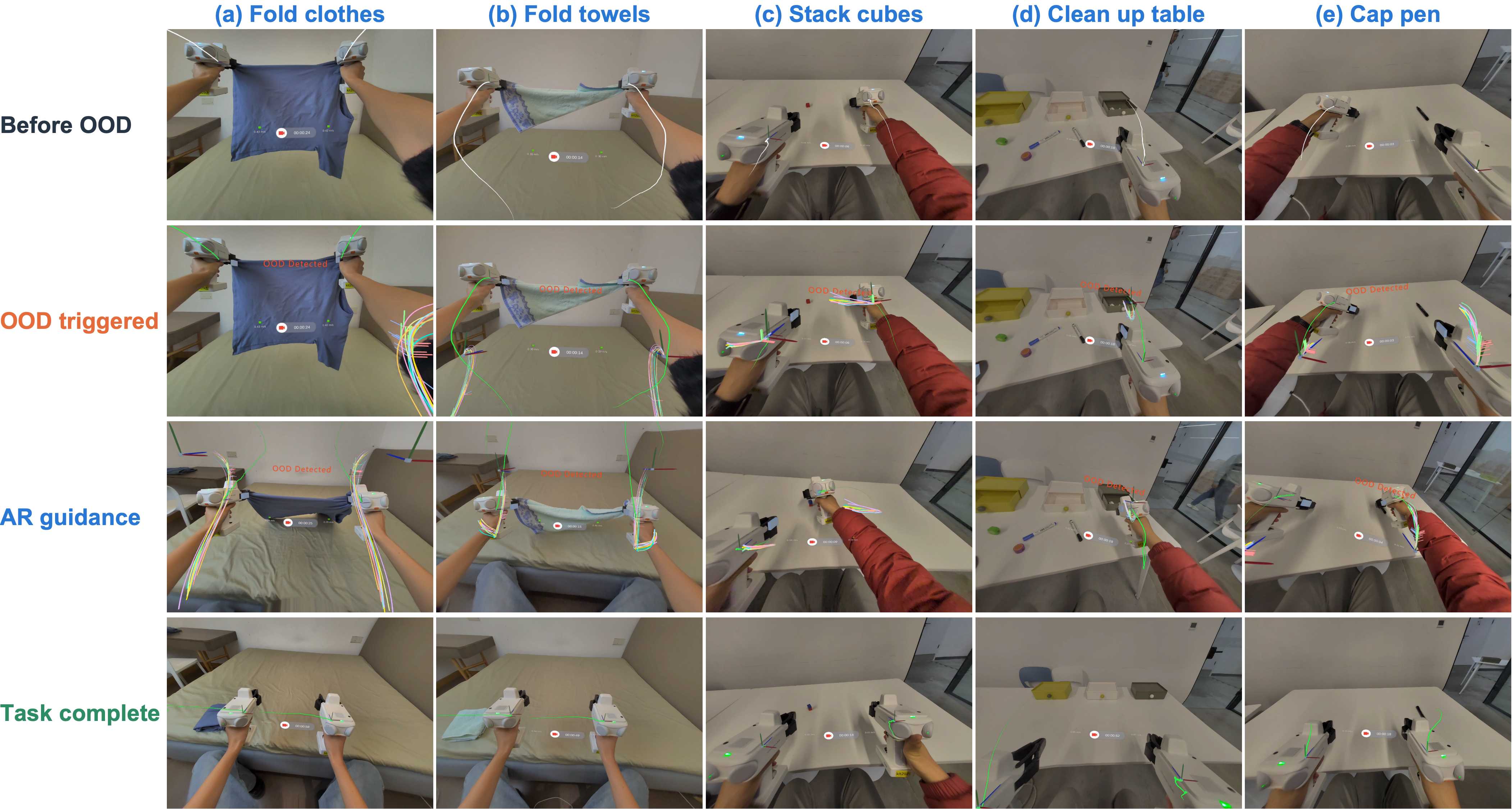}
\caption{\textbf{First-person OOD feedback and AR guidance across five tasks.}
After an OOD trigger, AR displays the end-effector coordinate frames at the
original observation time $t$, together with 10 policy-predicted trajectories
starting from those frames. The executed end-effector trajectories change
from white to green after the trigger for emphasis. OOD detection continues
until the first trigger; only then is further OOD evaluation disabled for the
remainder of the episode. The visualized trajectories subsequently disappear
gradually from the display within 4\,s. This limits visual clutter and avoids
distracting operators or disrupting their collection rhythm.}
\label{fig:supp_ood_ar_examples}
\end{figure}

\FloatBarrier
\paragraph{Displayed feedback versus observation time.}
The first visible alert marks feedback presentation in the recording, not
the observation time or the start of a cloud request. Under asynchronous
inference, the alert is displayed after predictions have returned and been
compared with buffered human actions.
The label is assigned to the original observation at time $t$, and its spatial
marker refers to the end-effector position at that observation time, not the
position at the trigger step $t+j^*$ or at prediction return.

\FloatBarrier
\section{Experiment Setup}
\label{ap:setup}
Table~\ref{tab:tasks} lists the task-specific base datasets, per-round training-episode quotas, and OOD thresholds. Both methods start from the same base model. For \sysname, only valid OOD episodes count toward the quota; training retains the $t$-to-end suffix. Quotas and thresholds vary by task difficulty.

\begin{table}[!htbp]
\centering
\small
\caption{Task settings. Data in trajectories (hours).}
\label{tab:tasks}
\setlength{\tabcolsep}{6pt}
\renewcommand{\arraystretch}{1.05}
\begin{tabularx}{\linewidth}{@{}l*{3}{>{\centering\arraybackslash}X}@{}}
\toprule
Task & Base data & Per round & Threshold \\
\midrule
Fold clothes   & 5,172 (60.0\,h) & 120 & 0.48 \\
Fold towels    & 1,630 (18.0\,h) & 100 & 0.22  \\
Stack cubes    & 1,954 (6.2\,h)  & 100 & 0.08  \\
Cap pen        & 987 (14.1\,h)  & 100  & 0.11  \\
Clean up table & 925 (6.6\,h)   & 50 & 0.18   \\
\bottomrule
\end{tabularx}
\end{table}

\clearpage
\section{Task Progress Score Rubric}
\label{sec:rubric}
In Table~\ref{tab:rubric}, we specify the score awarded for completing each milestone in every task. All scores reported in the main paper are calculated consistently according to these task-specific scoring criteria.
\par\medskip\noindent
\begin{minipage}{\linewidth}
\centering
\small
\captionof{table}{Milestones and scores for the Task Progress Score.}
\label{tab:rubric}
\setlength{\tabcolsep}{6pt}
\renewcommand{\arraystretch}{1.05}
\begin{tabularx}{\linewidth}{@{}p{0.23\linewidth}rX@{}}
\toprule
Task & Score & Milestone \\
\midrule
\multirow{6}{*}{Fold clothes}
 & 15\%  & Cloth moved to the folding area \\
 & 50\%  & Cloth unfolded without overlap \\
 & 70\%  & Cloth fully flattened \\
 & 80\%  & Half fold completed \\
 & 90\%  & Folded into a square \\
 & 100\% & Placed back in the storage area \\
\midrule
\multirow{6}{*}{Fold towels}
 & 15\%  & Towel moved to the folding area \\
 & 50\%  & Towel unfolded without overlap \\
 & 70\%  & Towel fully flattened \\
 & 80\%  & Half fold completed \\
 & 90\%  & Folded into a square \\
 & 100\% & Placed back in the storage area \\
\midrule
\multirow{4}{*}{Stack cubes}
 & 30\%  & Cube picked up \\
 & 50\%  & Cube moved above the target \\
 & 80\%  & Cube placed on the target \\
 & 100\% & Arm moved away \\
\midrule
\multirow{6}{*}{Clean up table}
 & 25\%  & Two type-1 items placed in box 1 \\
 & 33\%  & Box 1 closed \\
 & 55\%  & Two type-2 items placed in box 2 \\
 & 66\%  & Box 2 closed \\
 & 91\%  & Two type-3 items placed in box 3 \\
 & 100\% & Box 3 closed \\
\midrule
\multirow{5}{*}{Cap pen}
 & 20\%  & Pen picked up \\
 & 40\%  & Cap picked up \\
 & 60\%  & Pen and cap held horizontally \\
 & 80\%  & Pen and cap joined \\
 & 100\% & Capped pen placed on the table \\
\bottomrule
\end{tabularx}
\end{minipage}
\par\medskip

\clearpage
\section{Training and Detector Configuration}
\label{ap:configuration}
Table~\ref{tab:configuration} lists the training configurations for base
SFT and DAgger post-training, along with the OOD detector settings.

\par\medskip\noindent
\begin{minipage}{\linewidth}
\centering
\small
\captionof{table}{Training and detector configuration. Shared optimization
settings apply to both base SFT and DAgger.}
\label{tab:configuration}
\setlength{\tabcolsep}{6pt}
\renewcommand{\arraystretch}{1.05}
\begin{tabularx}{\linewidth}{@{}p{0.46\linewidth}XX@{}}
\toprule
Setting & Base SFT & DAgger \\
\midrule
Global batch size & 4096 & 512 \\
Peak learning rate & $2\!\times\!10^{-4}$ & $7.071\!\times\!10^{-5}$ \\
\midrule
\multicolumn{3}{@{}l}{\textbf{Shared model and optimization settings}} \\
Policy architecture & \multicolumn{2}{l@{}}{$\pi_{0.5}$} \\
Update strategy & \multicolumn{2}{l@{}}{Full-parameter fine-tuning} \\
Optimizer & \multicolumn{2}{l@{}}{AdamW} \\
AdamW $(\beta_1,\beta_2,\epsilon)$ & \multicolumn{2}{l@{}}{$(0.9,\,0.95,\,10^{-8})$} \\
Weight decay & \multicolumn{2}{l@{}}{$10^{-10}$} \\
Training precision & \multicolumn{2}{l@{}}{Bfloat16} \\
Learning-rate schedule & \multicolumn{2}{l@{}}{Cosine decay to $10^{-6}$} \\
Warm-up duration & \multicolumn{2}{l@{}}{1 epoch} \\
Action horizon $H$ & \multicolumn{2}{l@{}}{50} \\
\midrule
\multicolumn{3}{@{}l}{\textbf{DAgger data usage and checkpoint selection}} \\
Training epochs  & \multicolumn{2}{l@{}}{6} \\
Deployed checkpoint & \multicolumn{2}{l@{}}{Final (6th) epoch} \\
Historical : new OOD data & \multicolumn{2}{l@{}}{1:1 by sampled data volume} \\
New OOD training segment & \multicolumn{2}{l@{}}{$t$-to-episode-end suffix} \\
\midrule
\multicolumn{3}{@{}l}{\textbf{Online OOD evaluation}} \\
Stochastic action chunks $M$ & \multicolumn{2}{l@{}}{10} \\
Human-action recording rate & \multicolumn{2}{l@{}}{30 Hz} \\
Position / rotation / gripper & \multicolumn{2}{l@{}}{0.45 / 0.45 / 0.1 (weights)} \\
\bottomrule
\end{tabularx}
\end{minipage}
\par\medskip

\FloatBarrier

\section{Data Pipeline Details}
\label{app:data_pipeline}

This appendix supplements the cleaning, annotation, and quality-inspection
stages described in Sec.~\ref{sec:dagger}.

\FloatBarrier
\subsection{Upload and Preprocessing}
\label{app:pipeline_upload}
\label{app:pipeline_preprocessing}
Each episode retains a persistent identifier, its collection round, policy
version, and timestamp-aligned OOD labels. Uploads are resumable, and the
identifier prevents duplicate registration after retries. Processing starts
once all required streams have arrived and passed integrity checks.
Preprocessing aligns the headset and left/right UMI streams to the
egocentric-video timeline and refines wrist poses using redundant tracking
signals. Unrecoverable recordings are rejected. Processing runs episode by
episode in the background, with automatic job retries and versioned outputs.

\FloatBarrier
\subsection{Data Cleaning}
\label{app:pipeline_cleaning}
For both wrists, speed, acceleration, and jerk are computed from the refined
poses; jerk measures motion jitter. Episodes with sustained abnormal motion
or an excessive fraction of abnormal frames are rejected. Recording-integrity
checks also reject prolonged tracking loss, missing or undecodable streams,
out-of-order timestamps, substantial frame loss or duplication, and clock
drift. Visible motion must agree temporally with tracked motion. Black frames,
prolonged repeated images, and substantial blur are additional rejection
conditions. These checks target recording artifacts, not policy disagreement.

\FloatBarrier
\subsection{Data Annotation}
\label{app:pipeline_valid_frames}
\paragraph{Valid frames.}
Proprioceptive signals identify the active manipulation interval, excluding
idle periods at the beginning and end. Gripper opening/closing events are
verified against wrist motion; the span between the first and last verified
event is extended along sustained wrist motion to include reaching and final
placement. For slow corrections, the start criterion is relative to the
episode's wrist speed. Episodes without verified grasp events use wrist
motion alone. Figure~\ref{fig:pipeline_valid_frames} illustrates this process.
For new OOD episodes, valid-frame annotation removes idle frames within the
$t$-to-end suffix specified in the main text; it does not alter the OOD onset
or include pre-OOD data.

\begin{figure}[!htbp]
\centering
\includegraphics[width=0.92\linewidth]{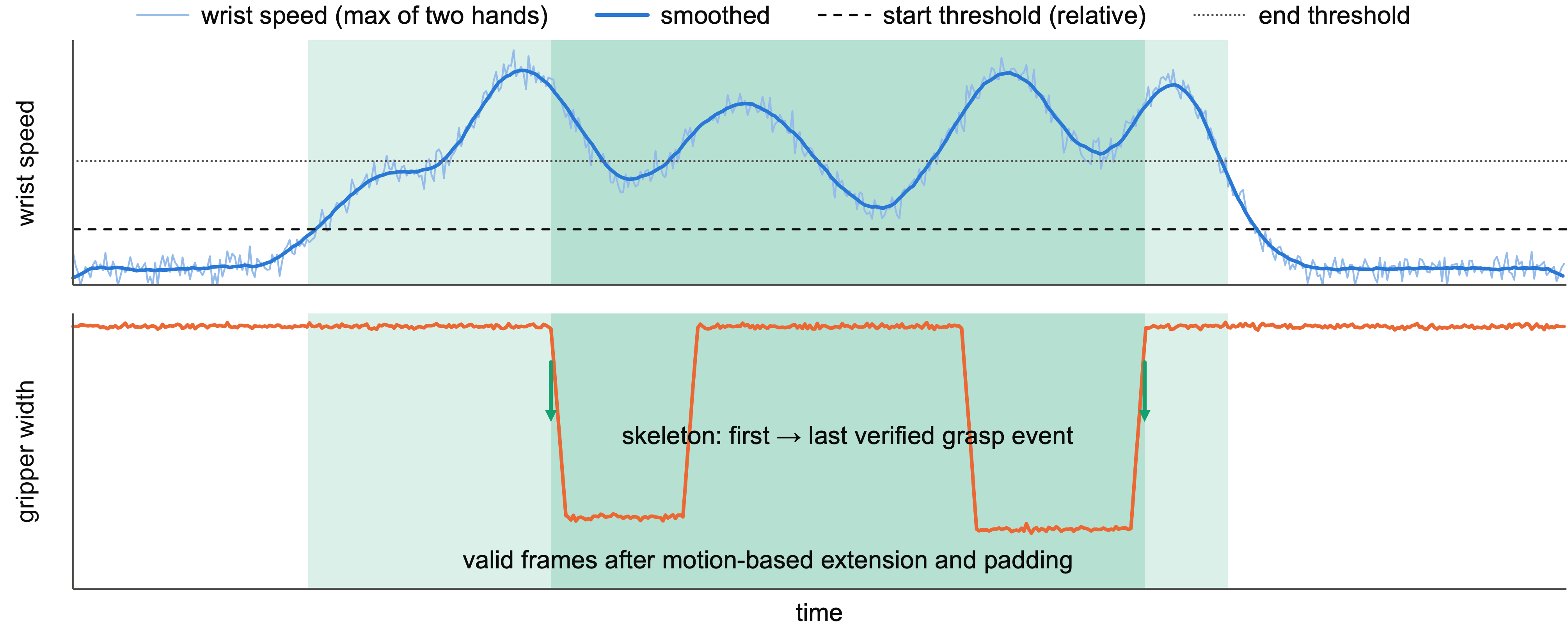}
\caption{\textbf{Valid-frame annotation on a schematic episode.}
\emph{Top:} raw and smoothed wrist speed (maximum over both hands) against the
relative start threshold and the end threshold.
\emph{Bottom:} gripper width. The dark band spans the first to the last verified
grasp event; the light band extends it along sustained wrist motion and adds
boundary padding, giving the valid frames. Traces and thresholds are
illustrative, not measurements from the five-task experiments.}
\label{fig:pipeline_valid_frames}
\end{figure}

\FloatBarrier

\paragraph{Semantic labels.}
Wrist and gripper signals propose action boundaries. A vision-language model
then examines the corresponding egocentric and wrist-camera spans to segment
atomic actions and assign subject--verb--object descriptions, such as
``left hand grasps cup.'' Boundaries are refined against state signals,
and overly short segments are merged with neighboring segments.

\FloatBarrier
\subsection{Quality Inspection}
\label{app:pipeline_quality}
The recorded videos are checked for task completion, visibility of grippers
and task-relevant objects, prolonged occlusion, compliance with the task's
required steps and placements, and consistency of the scene and objects.
Episodes failing these checks are excluded.

\FloatBarrier
\endgroup

\end{document}